\documentclass[journal]{IEEEtran}
\usepackage[backend=biber,sorting=none, style=ieee]{biblatex}
\usepackage{csquotes}
\usepackage[dvipsnames, table]{xcolor}
\definecolor{myblue}{HTML}{0068F0}
\definecolor{myorange}{HTML}{F08700}
\definecolor{myyellow}{HTML}{FFF28C}
\definecolor{mygreen}{HTML}{A9FF8C}
\usepackage[colorlinks=true, citecolor=myblue, linkcolor=myorange, urlcolor=myblue]{hyperref}
\usepackage[pdftex]{graphicx}
\usepackage{multirow}
\usepackage{makecell}
\def\hlinewd#1{%
\noalign{\ifnum0=`}\fi\hrule \@height #1 %
\futurelet\reserved@a\@xhline} 
\usepackage[cmex10]{amsmath}
\usepackage{amsfonts}
\usepackage{caption}
\usepackage{dirtytalk}
\usepackage{tikz}
\usepackage{hhline}
\usepackage{fixltx2e}
\usetikzlibrary{decorations.pathreplacing, calligraphy}
\newcommand{\tikzmark}[1]{\tikz[remember picture, overlay, baseline] \node[inner sep=0pt, outer sep=0pt] (#1) {};}

\begin{document}

\title{Benchmarking Hyperspectral Foundation Models for Hyperspectral Unmixing}

\author{Edgard~Dabier, Christophe~Kervazo, Pietro~Gori and Florence~Tupin\thanks{C. Kervazo would like to thank the ANR agency for the grant ANR-25-CE48-1042.}
         \\
        \textit{LTCI, Télécom Paris, Institut Polytechnique de Paris, Palaiseau, France}
}

\markboth{Journal of \LaTeX\ Class Files,~Vol.~13, No.~9, September~2014}%
{Shell \MakeLowercase{\textit{et al.}}: Benchmarking Hyperspectral Foundation Models for Hyperspectral Unmixing}

\maketitle

\begin{abstract}
Several foundation models dedicated to hyperspectral images have recently been made available. These models are trained on large unlabeled datasets and exhibit strong performance on many hyperspectral imaging tasks, such as classification or denoising. Nonetheless, their performance for hyperspectral unmixing -- the task of separating mixed spectra of overlapping materials in a hyperspectral image -- remain understudied. This might partly be due to the fact that most of them rely on vision transformer backbones, including patchification, leading to a feature resolution problem. While hyperspectral unmixing already arises from the low resolution of hyperspectral images, this patchification step potentially makes the problem even more ill-posed. Therefore, in this work, we aim to answer two questions: 1) \emph{how do foundation models perform in hyperspectral unmixing?}; 2) \emph{how to tackle the feature-level loss of resolution?} To answer the first question, we benchmark foundation models for unmixing, showing that they can reach state-of-the-art performance on four hyperspectral unmixing datasets. To answer the second question, we compare several feature upsampling approaches and empirically show that using a simple one can lead to high performance results. The code is available at \url{https://gitlab.telecom-paris.fr/ring/hfm-hsu.git}.
\end{abstract}

\begin{IEEEkeywords}
Blind Hyperspectral Unmixing, Remote Sensing, Hyperspectral Foundation Models, Linear Mixture Model.
\end{IEEEkeywords}

\IEEEpeerreviewmaketitle

\section{Introduction}

\IEEEPARstart{H}{yperspectral} images are composed of numerous narrow neighbouring spectral bands. Applied to remote sensing, this rich spectral content can be used to analyse the materials present in the scene, offering a vast set of applications in agriculture, mineral mapping, and space exploration \cite{Ren_2026}. Yet, due to physical constraints, hyperspectral images (HSI) have a poor spatial resolution: every pixel corresponds to a large area of the scene (\emph{e.g.} around 30 meters on the ground for satellite images) that likely contains more than one material. Thus, most observed pixels' spectra are a mixture of ``pure materials'' spectra, limiting the user's capacity to directly exploit the image. To alleviate this issue, hyperspectral unmixing (HSU) aims to decompose image pixels into a mixture of pure material spectra. In particular, blind HSU aims to estimate both the pure material spectra (endmembers) and their relative concentrations (abundances). \\

Due to its simplicity, the Linear Mixture Model (LMM), which assumes pixels to be a linear combination of the endmembers, is often used \cite{Ren_2026}. Mathematically, let $Y \in \mathbb{R}^{B, N}$ denote the matricised HSI with $N$ pixels ($N=H W$, $H$ and $W$ are the spatial dimensions) and $B$ spectral bands, $E \in \mathbb{R}^{B, C}$ the endmember matrix with $C$ pure materials and $A \in \mathbb{R}^{C, N}$ their corresponding abundance maps. The LMM can be written:

\vspace{-10pt}
\begin{equation}
    Y = EA + \varepsilon,
    \label{eq:unmix}
\end{equation}
\vspace{-15pt}

where $\varepsilon \in \mathbb{R}^{B, N}$ denotes additive noise. The LMM is often complemented by the abundance non-negativity constraint (ANC) and the abundance sum-to-one constraint (ASC).
 
\subsection{HSU state-of-the-art and motivation}

Due to their simplicity, pure pixel retrieval algorithms such as VCA \cite{Nascimento_2005} or N-FINDR \cite{Winter_1999} have quickly become popular in HSU. However, these approaches suffer from the fact that the pure-pixel assumption is rarely fulfilled in practice; optimisation-based methods leveraging more complex handcrafted priors such as minimum volume \cite{Heylen_2011} or sparsity \cite{Bioucas_2010} have thus taken the lion's share during the last decade. 

With the rise of deep learning, HSU has advanced significantly, particularly through convolutional architectures (CNNs), including CNNAEU \cite{Palsson_2021} and UnDIP \cite{Rasti_2022}. More recently, vision transformers (ViTs) such as DeepTrans \cite{Ghosh_2022} and A2SAN \cite{Tao_2024} have built on self-attention mechanisms to capture long-range spectral dependencies, while Mamba-based models like ProMU \cite{li_2025} have introduced efficient state-space modelling for spectral data. Additionally, adversarial schemes, including SSST-GAN \cite{Zhang_2026}, have been explored for blind HSU, learning underlying data distributions. Alternatively, some attempts have tried to bridge the gap between deep learning and optimisation approaches \cite{kervazo2026unrolled}.

Although these works have been reported to obtain accurate results, the networks are often trained based on an autoencoder-like framework over a \emph{single} image. This can be partly explained because manual HSU labelling requires considerable time and expertise, and therefore only a few scenes are annotated with the corresponding endmembers and abundances. Nonetheless, this is a big difference from many other computer vision tasks, where the networks often benefit from a vast amount of data. 

\subsection{Hyperspectral foundation models and contributions}

Despite this scarcity of annotated HSU images, researchers have recently gained access to vast unlabeled datasets of HSIs. This has enabled the development of Hyperspectral Foundation Models (HFMs), starting in 2024 with the seminal work of \cite{Hong_2024}. These HFMs, trained on large datasets, extract a representation of the input HSI, called features, that enables performing many downstream tasks without having to retrain the model for them.
Several HFMs were proposed: some of them use pixel-level feature encoders, such as HyperSIGMA \cite{Wang_2025}, HyperSL \cite{Kong_2025} or UniverSat \cite{perron_2026}. Other rather use patch-level features: HyperFree \cite{li_2025}, DOFA \cite{xiong_2025}, Panopticon \cite{waldmann_2025}, SpecViT (from  SpectralEarth \cite{Braham_2025}) and SpecAware \cite{Ji_2026}. Pixel-level encoders extract a representation of every pixel in the image, while patch-level encoders extract a representation of an entire patch, losing the fine details at the pixel scale.

In turn, patch-level encoders raise a specific issue for HSU: their features cannot be used directly, since the practitioner needs the proportion of each material at every pixel, not at the patch level. HyperFree \cite{li_2025}, the only paper using a patch-level HFM for HSU, does not explain how it handles this issue, and no code is available. This leaves a gap in the literature on how to use low-resolution feature HFMs for HSU.

More generally, only a limited number of results using HFMs for HSU were presented. The authors of \cite{Wang_2025} were the first to propose the use of HFM for HSU, but they only tested it on a single simple dataset. Later, HyperFree \cite{li_2025} reported comparable performance on the same dataset, and another article \cite{Liu_2025} suggested using HyperSL for HSU, but we couldn't reproduce the reported performance. Therefore, while there seems to be a rising interest in using HFMs for unmixing, existing literature still lacks a proper assessment of their performance over several datasets, with the associated codes for reproducible research.

In this context, the contributions of this work are the following:
\begin{itemize}
    \item we benchmark HFMs on the unmixing task over several datasets in order to better understand their performance in HSU. To this end, we build on a model-agnostic framework that performs unmixing directly from HFM features.
    \item We compare several feature upsampling strategies for patch-level HFM features.
\end{itemize}

\section{Methodology}

\begin{figure}
    \centering
    \includegraphics[width=\linewidth]{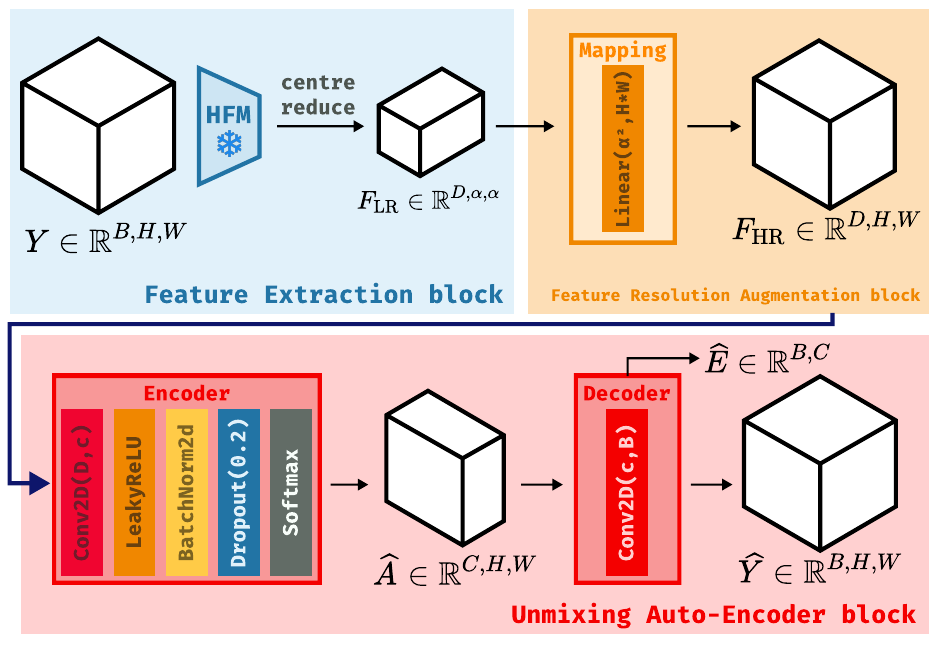}
    \caption{\small{Proposed HFM unmixing framework, divided into three blocks: the feature extraction block (forward pass through a frozen HFM, with $Y$ standardised following each model's specification), the feature resolution-augmentation block, and the unmixing auto-encoder block, based on a CNN auto-encoder architecture.}}
    \label{fig:fm-unmixing}
\end{figure}

To benchmark the existing HFMs' unmixing performance, we implement an unmixing model-agnostic framework, described in \autoref{sec:unmix_AE}, taking as input a resolution-augmented version of the HFMs' features. The resolution augmentation step, necessary to handle the loss of resolution in the ViT-based HFMs' feature space, is itself discussed in \autoref{features-mapping}. Please note that this step can be removed when using HFMs operating at pixel level (\emph{e.g.} \cite{perron_2026}). The training loss is described in subsection~\ref{sec:estimation-method}.
The whole framework is summarised in \autoref{fig:fm-unmixing}.

\subsection{Unmixing Auto-Encoder}
\label{sec:unmix_AE}
For the unmixing block, we use a classical auto-encoder structure, similar to the one in \cite{Palsson_2021}. We estimate abundances as latent vectors and use a linear decoder to mimic the LMM; its weights are then associated with the endmembers.

In detail, our unmixing block takes as input the resolution-augmented features derived from the considered HFM (see \autoref{features-mapping}) and aims to reconstruct the HSI.
The convolutional encoder reduces the spectral dimension of the high-resolution feature from $D$ to $C$ with a 2D CNN layer, followed by a LeakyReLU non-linearity, a 2D Batchnorm layer, a dropout layer with probability $0.2$ and finally a Softmax layer enforcing both the ANC and ASC constraints (\ref{eq:unmix}). The HSI is then reconstructed using a single matrix multiplication, increasing the spectral dimension from $C$ back to $B$. The decoder weights are initialised with a first guess of the endmembers obtained using the SiVM algorithm \cite{Heylen_2011}.

\subsection{Resolution augmentation of the HFM features}\label{features-mapping}

Most vision transformer-based foundation models extract features at the patch level in the input image. This means that the HSI is patchified with a patch size $p$, resulting in features $F_{\text{LR}} \in \mathbb{R}^{D \times \alpha \times \alpha}$ where $\alpha = \lfloor \frac{H}{p} \rfloor$, and $D$ is the embedding dimension. To obtain pixel-wise abundance estimation, the resolution of this feature map has to be augmented back to image resolution $F_{\text{HR}} \in \mathbb{R}^{D, H, W}$. We have tested several methods covering a wide range of approaches for this task. \\

We explored several widespread image super-resolution methods, including a non-learned method using a \emph{bilinear} upsampling (as in HyperSigma \cite{Wang_2025}), which interpolates the four nearest neighbour pixels. We also tested learned convolutional architectures: 1) \emph{deconvolution} reverses the roles of input/output in convolution by inserting zeros and applying transposed convolutions; 2) \emph{Pixel Shuffle} \cite{Shi_2016} expands channels via stride-1 convolution and rearranges them spatially; 3) the \emph{FiLM} feature fusion method \cite{Perez_2018}, which combines bilinear upsampling of low-resolution features with high-resolution HSI details predicted by a series of trainable 2D convolutions: $F_{\text{HR}}= f_1(\text{up}(F_{\text{LR}}) \cdot f_2(Y)+f_3(Y))$. Here, $f_1$ is a depthwise 2D convolutional layer with a kernel size of 3, $D$ input and output channels and a padding of 1; $f_2$ and $f_3$ are 2D convolutional layers with a kernel size of 1, $B$ input channels and $D$ output channels; and $\text{up}$ denotes the bilinear upsampling function.

Directly related to our goal, we also experimented with feature-upsampling networks, such as \emph{FeatUp} \cite{Fu_2024}, which is trained to enforce consistency across low-resolution feature maps under small image transformations, and \emph{NAF} \cite{Chambon_2025}, an attention-based method we trained on the SpectralEarth dataset to reconstruct high-resolution features.

Finally, we also tested a trained \emph{linear} mapping approach (as in \cite{Ghosh_2022}), which learns a direct transformation $W_{\text{up}} \in \mathbb{R}^{\alpha^2,HW}$ from low- to high-resolution feature spaces. It can be written as $F_{\text{HR}} = W_{\text{up}}F_{\text{LR}}$, with $W_{\text{up}}$ learned. We will show in \autoref{mapping} that this approach experimentally corresponds to the best strategy.

\subsection{Ensemble Learning Trick}
\label{sec:ensembling}

We experimentally observed that a simple way to obtain better performance is to use an ensemble learning method \cite{Montgomery_2012}. We run $n$ independent training runs of the described framework to obtain $n$ estimates of $\widehat{A}_i$ and $\widehat{E}_i$, and retrieve the final unmixing prediction by averaging these individual predictions:
\vspace{-10pt}
\begin{equation}
    \widehat{A} = \displaystyle \frac{1}{n} \sum_{i=0}^n\widehat{A}_i \text{, and } \widehat{E} = \displaystyle \frac{1}{n} \sum_{i=0}^n\widehat{E}_i.
    \label{eq:ensembling}
\end{equation}
\vspace{-10pt}

This improves the estimation quality while reducing the variability of the model, at the cost of a slight increase in computation time. 

\vspace{-10pt}
\subsection{Training loss}
\label{sec:estimation-method}

The proposed unmixing framework is trained end-to-end in an unsupervised manner by minimising a mixture of losses :

\vspace{-5pt}
\begin{equation}
    \text{SAD}(\widehat{Y}, Y) + \lambda_1 \text{NMSE}(\widehat{Y}, Y) + \lambda_2 \mathcal{L}_{\frac{1}{2}}(\widehat{A}),
\end{equation}

with for any dumb variable matrices $X,\widehat{X} \in \mathbb{R}^{D_1,D_2}$:
\small
\begin{equation}
    \text{SAD}(\widehat{X}, X) = \frac{1}{D_1} \sum_{i=1}^{D_1} \arccos \left( \sum_{j=1}^{D_2} \frac{X_{j,i}}{\| \mathbf{X}_{:,i} \|_2} \cdot \frac{\widehat{X}_{j,i}}{\| \widehat{\mathbf{X}}_{:,i} \|_2} \right)
    \label{eq:sad}
\end{equation}
\normalsize
\begin{equation}
    \text{NMSE}(\widehat{X}, X) = \displaystyle \frac{\sum_{i=1}^{D_1} \sum_{j=1}^{D_2} (X_{i,j} - \widehat{X}_{i,j})^2}{\sum_{i=1}^{D_1} \sum_{j=1}^{D_2} (X_{i,j})^2}
    \label{eq:nmse}
\end{equation}
\begin{equation}
    \mathcal{L}_{\frac{1}{2}}(\widehat{A}) = \displaystyle \frac{1}{CN} \sum_{i=0}^{C} \sum_{j=0}^N\sqrt{\widehat{A}_{i,j}},
    \label{eq:reg}
\end{equation}

where ${\text{SAD}}(\widehat{Y}, Y)$ is known as the Spectral Angular Distance (\ref{eq:sad}) and ${\text{NMSE}}(\widehat{Y}, Y)$ is the Normalized Mean Squared Error \eqref{eq:nmse} (with reduction set to \emph{sum}), assessing the quality of the reconstruction. In addition, the $\frac{1}{2}$-norm regularisation term encourages sparsity in the abundance estimation (\ref{eq:reg}). 

\section{Experimental Results}

\subsection{Data Description}

Experiments were conducted on several available labelled datasets, providing an in-depth comparison of the methods :
\begin{itemize}
    \item Samson: a relatively simple dataset, with a spatial size of $95\times 95$, $156$ spectral bands, and 3 endmembers.
    \item Jasper Ridge: it has $100 \times 100$ pixels, $198$ spectral bands and 4 endmembers.
    \item Apex: has $110 \times 110$ pixels, $285$ spectral bands and contains 4 endmembers.
    \item Urban: has a significantly larger spatial extent of $307$ pixels, 162 spectral bands and 6 endmembers
\end{itemize}

Since most HFM have a fixed input size (usually $224 \times 224$), we upsampled the smaller images (using bilinear upsampling) and cropped Urban, which is larger. This preprocessing was applied to all methods for fair comparison.

\subsection{Experimental Setup}\label{sec:setup}

The estimated endmember quality is assessed through the SAD \eqref{eq:sad} and the abundance through the NMSE (\ref{eq:nmse}). Both are averaged over all endmembers. We also compare the execution time of all methods (measured on an NVIDIA V100 GPU). For easier readability, we compute a score $s_i$ for each method $i$, aggregating over all the datasets $j\in \{1..J\}$ (here we consider $J=4$ datasets) the SAD and NMSE after normalisation on each dataset $j$ and average this score to obtain a global ranking of the methods (\emph{Rank}). The score is obtained with:
\begin{equation}
    s_i = \displaystyle  \frac{1}{2J}\sum_{j=1}^{J} \left( \frac{\text{SAD}_{i,j}}{\text{max}_j(\text{SAD}_{i,j})}+\frac{\text{NMSE}_{i,j}}{\text{max}_j(\text{NMSE}_{i,j})}\right), 
\end{equation}
where $\text{SAD}_{i,j}$ and $\text{NMSE}_{i,j}$ are the metrics of method $j$ on dataset $i$. This score is used to obtain a global ranking of the methods (\emph{Rank}). The training and loss hyperparameters are chosen to be the same across all the datasets. They were obtained using the Optuna Bayesian optimisation framework \cite{optuna_2019}, resulting in a weight decay of $0.18$, a learning rate of $2e-3$ and $\lambda_1=0.09$ and $\lambda_2=0.6$. We also used $200$ training epochs and the AdamW optimiser.

\subsection{Choice of the Number of Trainings $n$ for Ensembling}

\begin{figure}[h!]
    \centering
    \begin{minipage}{0.49\linewidth}
        \centering
        \includegraphics[width=\linewidth]{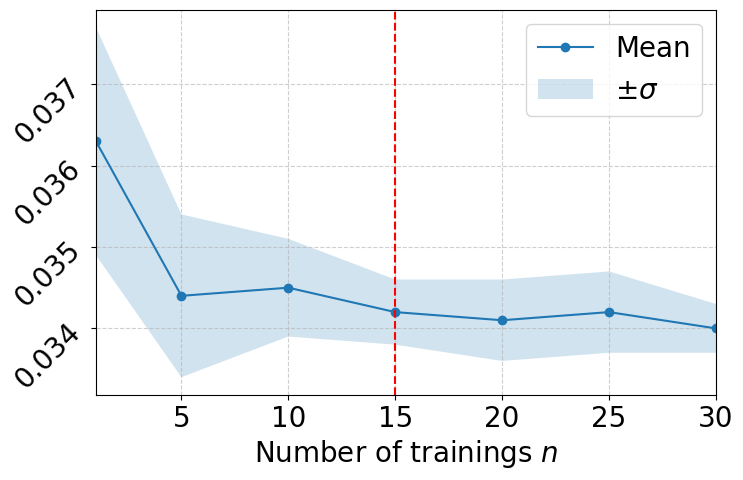} \\
        Samson
    \end{minipage}
    \hfill
    \begin{minipage}{0.49\linewidth}
        \centering
        \includegraphics[width=\linewidth]{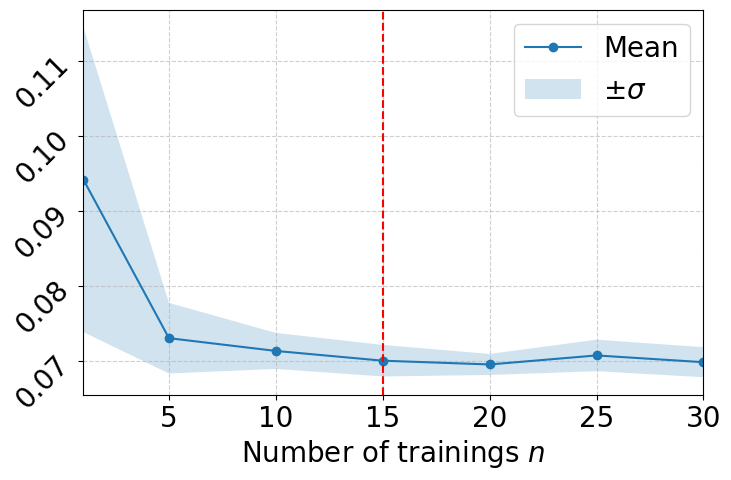} \\
        Jasper
    \end{minipage}
    \caption{\small{Evolution of the mean of SAD and NMSE as a function of $n$, the number of training runs, for two datasets. The plain line is the average score for 10 runs of $n$ trainings, and the blue area is the associated std.}}
    \label{fig:n-training} 
\end{figure}

We study here the impact of the number of training runs $n$ in the ensembling method. \autoref{fig:n-training} displays the evolution of the metrics as a function of $n$, using the Panopticon HFM. To better assess the robustness of the method, the average over 10 experiments is displayed (solid line), as well as the associated standard deviation. As can be seen, ensemble clearly improves the results while reducing the variability. In practice, for $ n> 15$, there is no further improvement. 

\vspace{-10pt}
\subsection{Assessing of the Feature Resolution Augmentation strategy}\label{mapping}

Feature resolution augmentation plays an essential role in the method, and we tested several strategies for it (see \autoref{features-mapping}). \autoref{tab:upsampling-methods-15trains} compares the resolution augmentation methods \emph{in terms of the final unmixing quality} they lead to, once integrated in our unmixing pipeline. The low-resolution features are obtained by the Panopticon HFM (but we obtained similar results with other HFMs - see supplementary material) with a 15-training ensembling (denoted by \textit{($\times15$)} in the table; see \autoref{sec:ensembling}).
It also shows the average of the metrics over all datasets for easier comparison (\textit{Avg.}), as well as the number of trainable parameters of each method (\textit{Size}).

\begin{table}[h]
    \centering
    \renewcommand{\arraystretch}{1.2}
    \addtolength{\tabcolsep}{-4pt}
    \begin{tabular}{|l|c|c|c|c|c|c|c|c|}
        \hline
        Method & Samson & Jasper & Apex & Urban & Avg. & Size \\
        \Xhline{4\arrayrulewidth}
        Bilinear ($\times 15$) & 0.054 & 0.164 & 0.352 & 0.321 & 0.222 & 0 \\
        \hline
        Deconvolution ($\times 15$) & \textcolor{myorange}{\underline{0.038}} & \textcolor{myorange}{\underline{0.098}} & 0.157 & \textcolor{myorange}{\underline{0.169}} & \textcolor{myorange}{\underline{0.115}} & 263k \\
        \hline
        Pixel Shuffle ($\times 15$) \cite{Shi_2016} & 0.057 & 0.201 & 0.630 & 0.347 & 0.308 & 524k \\
        \hline
        FiLM layer ($\times 15$) \cite{Perez_2018} & 0.04 & 0.123 & \textcolor{myblue}{\textbf{0.140}} & 0.184 & 0.122 & 331k \\
        \hline
        NAF ($\times 15$) \cite{Chambon_2025} & 0.480 & 0.485 & 0.429 & 0.545 & 0.485 & 879k \\
        \hline
        FeatUp ($\times 15$) \cite{Fu_2024} & 0.615 & 0.719 & 0.434 & 0.646 & 0.603 & 668k \\
        \Xhline{4\arrayrulewidth}
        \textbf{Linear Mapping ($\times 15$)} \cite{Ghosh_2022} & \textcolor{myblue}{\textbf{0.036}} & \textcolor{myblue}{\textbf{0.083}} & \textcolor{myorange}{\underline{0.167}} & \textcolor{myblue}{\textbf{0.151}} & \textcolor{myblue}{\textbf{0.109}} & 9.8M \\
        \hline
    \end{tabular}
    \caption{Comparison of upsampling strategies' performance.}
    \label{tab:upsampling-methods-15trains}
\end{table}

The linear mapping strategy $(\times 15)$ appears to be the best choice, outperforming the others on almost all datasets.
In the case of limited resources (in particular when it is not possible to perform $n=15$ training runs for ensembling), we show in the supplementary materials that FiLM might be a good alternative.
Surprisingly, the two methods which were designed for feature super-resolution, NAF and FeatUp, obtain the two worst results. 

\subsection{HFM benchmark for HSU}
\begin{figure*}[h]
    \includegraphics[width=\linewidth]{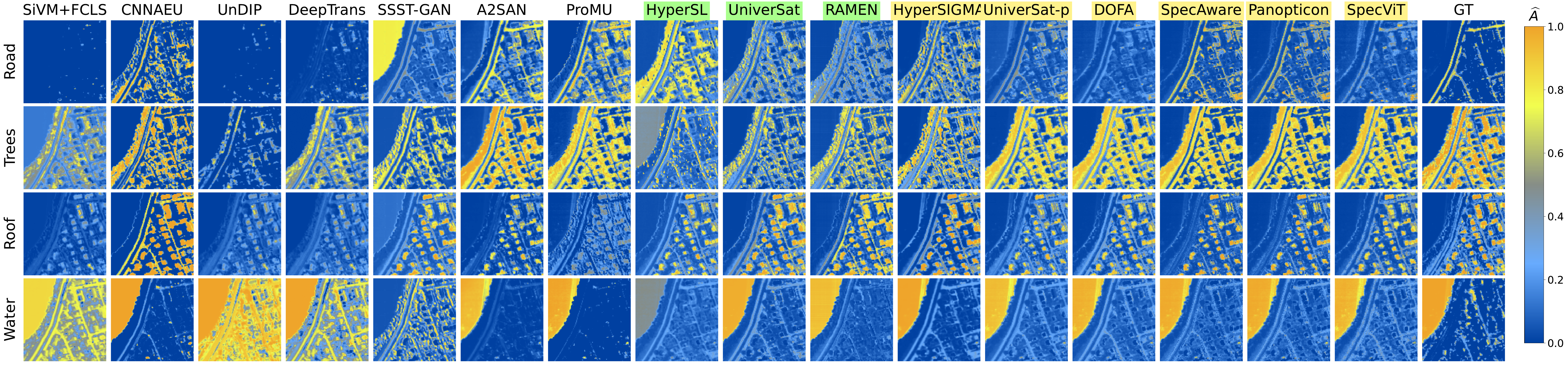}
    \caption{Comparison of all method's estimated $\widehat{A}$ on the Apex dataset, highlights show \colorbox{mygreen}{pixel-level} and \colorbox{myyellow}{path-level} HFMs.}
    \label{fig:a-comparison}
\end{figure*}

To evaluate our proposed framework, we compare the averaged scores obtained on 10 independent training runs with 7 state-of-the-art methods. These methods span a diverse range of approaches, from classic optimisation methods using SiVM and FCLS \cite{Heylen_2011}\cite{Heinz_1999}, to CNNs with CNNAEU \cite{Palsson_2021} and UnDIP \cite{Rasti_2022}, ViTs with DeepTrans \cite{Rasti_2022}, A2SAN \cite{Tao_2024}, GANs with SSST-GAN \cite{Zhang_2026}, and Mamba with ProMU \cite{Liu_2025} (we could not get results on the Urban dataset using the available code). To these methods, we compare all the previously mentioned HFMs \footnote{For HyperSL, we had to crop the Urban dataset to size $128 \times 128$ for memory issues; extracting features from images of size $224 \times 224$ uses more than 100 GB VRAM, and HyperFree's available pretrained weights were missing the positional embedding, making it impossible to use.}, once integrated into our pipeline. 
\autoref{tab:methods-comparison} displays the results obtained with DOFA (DOFA v1-large, as we observed better results than with v2), Panopticon, SpecViT-b, SpecAware and UniverSat using the proposed framework (without the mapping block for UniverSat, as it is a pixel-level encoder). We also provide results obtained with ($n=15$) and without ($n=1$) the ensembling method.
Qualitative results about the estimation of abundances on the Apex dataset are shown in \autoref{fig:a-comparison}.

\begin{table*}[h]
    \centering
    \addtolength{\tabcolsep}{-3.5pt}
    \renewcommand{\arraystretch}{1.2}
    \begin{tabular}{|l|c|c|c|c|c|c|c|c|c|c|c|c|c|c|}
    \hline
    \multirow{2}{*}{} &
    \multicolumn{3}{c|}{\textbf{Samson}} &
    \multicolumn{3}{c|}{\textbf{Jasper}} &
    \multicolumn{3}{c|}{\textbf{Apex}} &
    \multicolumn{3}{c|}{\textbf{Urban}} & \multirow{2}{*}{Score} &  \multirow{2}{*}{Rank} \\
    \cline{2-13}
    & \textbf{SAD $\downarrow$} & \textbf{NMSE $\downarrow$} & \textbf{Time} & \textbf{SAD $\downarrow$} & \textbf{NMSE $\downarrow$} & \textbf{Time} & \textbf{SAD $\downarrow$} & \textbf{NMSE $\downarrow$} & \textbf{Time} & \textbf{SAD $\downarrow$} & \textbf{NMSE $\downarrow$} & \textbf{Time} & & \\
    \Xhline{4\arrayrulewidth}
    \tikzmark{r1} SiVM + FCLS \cite{Heinz_1999} & 0.056\textsubscript{± 0} & 0.438\textsubscript{± 0} & 1.8 & 0.267\textsubscript{± 0} & 0.187\textsubscript{± 0} & 2.4 &  0.244\textsubscript{± 0} & 0.938\textsubscript{± 0} & 2.4 &  0.394\textsubscript{± 0} & 0.870\textsubscript{± 0} & 2.4 & 5.4 & 20 \\
    \hline
    \tikzmark{r2} CNNAEU \cite{Palsson_2021} & 0.086\textsubscript{± 0.09} & 0.114\textsubscript{± 0.11} & 24 &  0.166\textsubscript{± 0.12} & 0.277\textsubscript{± 0.10} & 30 &  0.301\textsubscript{± 0.23} & 0.366\textsubscript{± 0.18} & 42 &  0.175\textsubscript{± 0.06} & 0.464\textsubscript{± 0.16} & 36 & 4.2 & 17 \\
    \hline
    \tikzmark{r3} UnDIP \cite{Rasti_2022} & 0.056\textsubscript{± 0} & 0.440\textsubscript{± 0.03} & 132 & 0.267\textsubscript{± 0} & 0.191\textsubscript{± 0.01} & 144 & 0.244\textsubscript{± 0} & 0.991\textsubscript{± 0.04} & 270 & 0.394\textsubscript{± 0} & 0.866\textsubscript{± 0} & 138 & 5.5 & 21 \\
    \hline
    \tikzmark{r4} DeepTrans \cite{Ghosh_2022} & 0.205\textsubscript{± 0} & 0.343\textsubscript{± 0} & 18 & 0.266\textsubscript{± 0.01} & 0.256\textsubscript{± 0.01} & 24 & 0.288\textsubscript{± 0} & 0.428\textsubscript{± 0.01} & 24 & 0.456\textsubscript{± 0.01} & 0.515\textsubscript{± 0.02} & 36 & 5.3 & 19 \\
    \hline
    \tikzmark{r5} SSST-GAN \cite{Zhang_2026} & 0.138\textsubscript{± 0.01} & 0.082\textsubscript{± 0.02} & 150 & 0.261\textsubscript{± 0} & 0.181\textsubscript{± 0.08} & 180 & 0.405\textsubscript{± 0.08} & 0.832\textsubscript{± 0.04} & 258 & 0.260\textsubscript{± 0.01} & 0.659\textsubscript{± 0.09} & 180 & 5.1 & 18 \\
    \hline
    \tikzmark{r6} A2SAN \cite{Tao_2024} & 0.061\textsubscript{± 0.01} & \textcolor{myblue}{\textbf{0.014\textsubscript{± 0}}} & 18 & 0.139\textsubscript{± 0.02} & 0.079\textsubscript{± 0.03} & 16 & 0.198\textsubscript{± 0.19} & 0.202\textsubscript{± 0.14} & 17 & 0.203\textsubscript{± 0.05} & 0.346\textsubscript{± 0.09} & 17 & 3.1 & 12 \\
    \hline
    \tikzmark{r7} ProMU \cite{Liu_2025} & 0.135\textsubscript{± 0.01} & 0.039\textsubscript{± 0} & 3400 & 0.185\textsubscript{± 0.03} & 0.195\textsubscript{± 0.10} & 852 & 0.151\textsubscript{± 0} & 0.250\textsubscript{± 0.01} & 1170 & // & // & // & 3.5 & 14 \\
    \Xhline{3\arrayrulewidth}
    \tikzmark{r8} HyperSL \cite{Kong_2025} & 0.107\textsubscript{± 0} & 0.044\textsubscript{± 0} & 1044 & 0.155\textsubscript{± 0} & \textcolor{myorange}{\underline{0.078\textsubscript{± 0.01}}} & 1400 & 0.320\textsubscript{± 0.21} & 0.209\textsubscript{± 0.11} & 2100 & 0.326\textsubscript{± 0.01} & 0.518\textsubscript{± 0.06} & 1092 & 3.8 & 16 \\
    \hline
    \tikzmark{r9} UniverSat \cite{perron_2026} & 0.094\textsubscript{± 0.04} & 0.106\textsubscript{± 0.05} & 645 & 0.104\textsubscript{± 0.01} & 0.131\textsubscript{± 0.01} & 838 &  0.555\textsubscript{± 0} & 0.363\textsubscript{± 0.02} & 556 & 0.193\textsubscript{± 0.01} & 0.206\textsubscript{± 0.01} & 666 & 3.7 & 15 \\
    \hline
    \tikzmark{r10} RAMEN \cite{Houdre_2026} & 0.039\textsubscript{± 0} & 0.036\textsubscript{± 0} & 28 & 0.104\textsubscript{± 0.01} & 0.117\textsubscript{± 0.03} & 28 & \textcolor{myorange}{\underline{0.140\textsubscript{± 0.01}}} & 0.189\textsubscript{± 0.02} & 28 & 0.188\textsubscript{± 0.02} & 0.319\textsubscript{± 0.03} & 28 & 2.7 & 10 \\
    \hline
    \tikzmark{r11} HyperSIGMA \cite{Wang_2025} & \textcolor{myblue}{\textbf{0.027\textsubscript{± 0}}} & \textcolor{myorange}{\underline{0.025\textsubscript{± 0}}} & 4300 &  0.146\textsubscript{± 0} & 0.280\textsubscript{± 0} & 4400 & 0.462\textsubscript{± 0} &  0.402\textsubscript{± 0} & 4440 & 0.221\textsubscript{± 0} & 0.213\textsubscript{± 0} & 4400 & 3.2 & 13 \\
    \hline
    \tikzmark{r12} UniverSat-p \cite{perron_2026} & 0.033\textsubscript{± 0} & 0.043\textsubscript{± 0} & 75 & 0.099\textsubscript{± 0.01} & 0.092\textsubscript{± 0.02} & 85 & 0.160\textsubscript{± 0.12} & \textcolor{myorange}{\underline{0.137\textsubscript{± 0.01}}} & 133 & 0.189\textsubscript{± 0.02} & 0.162\textsubscript{± 0.01} & 300 & 1.8 & 6 \\
    \hline
    \tikzmark{r13} DOFA \cite{xiong_2025} & 0.033\textsubscript{± 0} & 0.044\textsubscript{± 0.01} & 16 & 0.124\textsubscript{± 0.08} & 0.129\textsubscript{± 0.13} & 17 & 0.220\textsubscript{± 0.17} & 0.219\textsubscript{± 0.15} & 20 & 0.182\textsubscript{± 0.03} & 0.232\textsubscript{± 0.15} & 16 & 3 & 11 \\
    \hline
    \tikzmark{r14} SpecAware \cite{Ji_2026} & \textcolor{myorange}{\underline{0.028\textsubscript{± 0}}} & 0.043\textsubscript{± 0} & 19 & 0.080\textsubscript{± 0} & 0.099\textsubscript{± 0.03} & 19 & 0.172\textsubscript{± 0.17} & 0.255\textsubscript{± 0.19} & 20 & \textcolor{myorange}{\underline{0.149\textsubscript{± 0.02}}} & 0.168\textsubscript{± 0.01} & 19 & 1.5 & 5 \\
    \hline
    \tikzmark{r15} Panopticon \cite{waldmann_2025} & 0.029\textsubscript{± 0} & 0.046\textsubscript{± 0} & 63 & 0.084\textsubscript{± 0.01} & 0.111\textsubscript{± 0.03} & 78 & 0.294\textsubscript{± 0.21} & 0.418\textsubscript{± 0.28} & 111 & 0.157\textsubscript{± 0.02} & 0.173\textsubscript{± 0.01} & 65 & 2.5 & 8 \\
    \hline
    \tikzmark{r16} SpecViT \cite{Braham_2025} & 0.029\textsubscript{± 0} & 0.044\textsubscript{± 0} & 57 & 0.090\textsubscript{± 0.05} & 0.124\textsubscript{± 0.04} & 58 & 0.210\textsubscript{± 0.19} & 0.317\textsubscript{± 0.26} & 57 & 0.196\textsubscript{± 0} & 0.268\textsubscript{± 0.01} & 57 & 2.6 & 9 \\
    \Xhline{3\arrayrulewidth}
    \tikzmark{r17} DOFA \textit{($\times$15)} & 0.033\textsubscript{± 0} & 0.043\textsubscript{± 0} & 261 & 0.094\textsubscript{± 0} & \textcolor{myorange}{\underline{0.078\textsubscript{± 0.01}}} & 261 & 0.212\textsubscript{± 0.11} & 0.212\textsubscript{± 0.06} & 300 & 0.159\textsubscript{± 0.01} & 0.150\textsubscript{± 0.01} & 251 & 2 & 7 \\
    \hline
    \tikzmark{r18} SpecAware \textit{($\times$15)} & 0.029\textsubscript{± 0} & 0.040\textsubscript{± 0} & 260 & \textcolor{myorange}{\underline{0.077\textsubscript{± 0}}} & \textcolor{myblue}{\textbf{0.060\textsubscript{± 0}}} & 262 & 0.157\textsubscript{± 0.06} & 0.173\textsubscript{± 0.02} & 268 &  \textcolor{myblue}{\textbf{0.147\textsubscript{± 0.01}}} & 0.149\textsubscript{± 0} & 263 & 0.8 & \textcolor{myblue}{\textbf{1}} \\
    \hline
    \tikzmark{r19} Panopticon \textit{($\times$15)} & 0.029\textsubscript{± 0} & 0.042\textsubscript{± 0} & 920 & 0.087\textsubscript{± 0} & \textcolor{myorange}{\underline{0.078\textsubscript{± 0}}} & 1150 & 0.168\textsubscript{± 0.03} & 0.166\textsubscript{± 0.01} & 1650 & 0.154\textsubscript{± 0} & \textcolor{myorange}{\underline{0.148\textsubscript{± 0}}} & 966 & 1.1 & \textcolor{myorange}{\underline{2}}\\
    \hline
    \tikzmark{r20} SpecViT \textit{($\times$15)} & 0.029\textsubscript{± 0} & 0.040\textsubscript{± 0} & 850 & \textcolor{myblue}{\textbf{0.075\textsubscript{± 0}}} & \textcolor{myblue}{\textbf{0.06\textsubscript{± 0}}} & 858 & 0.188\textsubscript{± 0.11} & 0.191\textsubscript{± 0.05} & 864 & 0.195\textsubscript{± 0} & 0.255\textsubscript{± 0} & 858 & 1.47 & 3 \\
    \hline
    \tikzmark{r20} UniverSat-p \textit{($\times$15)} & 0.032\textsubscript{± 0} & 0.042\textsubscript{± 0} & 1080 & 0.096\textsubscript{± 0} & 0.083\textsubscript{± 0.01} & 1260 & \textcolor{myblue}{\textbf{0.123\textsubscript{± 0.03}}} & \textcolor{myblue}{\textbf{0.125\textsubscript{± 0}}} & 1800 & 0.182\textsubscript{± 0.01} & \textcolor{myblue}{\textbf{0.140\textsubscript{± 0}}} & 1800 & 1.48 & 4 \\
    \hline
    \end{tabular}
    \caption{\normalsize{Comparison of all HSU methods on the 4 datasets, averaged over 10 runs, best results in \textcolor{myblue}{\textbf{bold}}, second best \textcolor{myorange}{\underline{underlined}}. UniverSat is tested with pixel-level and patch-level (\emph{UniverSat-p}) feature extraction. \emph{pixel-level} and \emph{patch-level} gather HFM extracting pixel- and patch-level features, respectively, and \emph{ensembling} gathers results obtained with the ensembling method. We display results as Mean\textsubscript{± std}, \emph{Time} is in seconds, and \emph{Rank} is obtained with the score described in \autoref{sec:setup}.}}

    \begin{tikzpicture}[remember picture, overlay]
        \draw[decorate, decoration={brace, amplitude=3pt, mirror, raise=3pt}, thick]
            (r1.north west) -- (r7.south west)
            node[midway, xshift=-11pt, rotate=90, anchor=center] {HSU SOTA};
        
        \draw[decorate, decoration={brace, amplitude=3pt, mirror, raise=3pt}, thick]
            (r8.north west) -- (r10.south west)
            node[midway, xshift=-11pt, rotate=90, anchor=center] {pixel-level};
            
        \draw[decorate, decoration={brace, amplitude=3pt, mirror, raise=3pt}, thick]
            (r11.north west) -- (r16.south west)
            node[midway, xshift=-11pt, rotate=90, anchor=center] {patch-level};
        
        \draw[decorate, decoration={brace, amplitude=3pt, mirror, raise=3pt}, thick]
            (r17.north west) -- (r20.south west)
            node[midway, xshift=-11pt, rotate=90, anchor=center] {ensembling};
    \end{tikzpicture}
    
    \label{tab:methods-comparison}
\end{table*}

We observe that HFMs can achieve results comparable to the state-of-the-art in HSU. Moreover, we observe better performance with models that extract low-resolution features than with pixel-level encoders, highlighting the value of working with ViTs, even if they introduce a spatial downsampling. Finally, we note that the ensembling method enhances the quality of the prediction, even if a single training can already achieve decent performance. We note that the computational overhead of doing 15 trainings remains smaller than a single training of the unmixing part of HyperSIGMA or HyperSL.

\section{Conclusion}

In this letter, we showcased the ability of foundation models to perform efficient unmixing by introducing a simple-to-implement,  model-agnostic method allowing us to benchmark HFMs on the HSU task. We showed that they can achieve state-of-the-art performance while benchmarking different potential solutions to the resolution loss introduced by ViT-based models. 

\printbibliography

\newpage

\appendices

\section{Choice of the Feature Mapping}

\begin{table*}
    \centering\renewcommand{\arraystretch}{1.2}
    \begin{tabular}{|l|c|c|c|c|c|c|c|c|c|c|}
        \hline
        \multirow{2}{*}{} &
        \multicolumn{2}{c|}{\textbf{Samson}} &
        \multicolumn{2}{c|}{\textbf{Jasper}} &
        \multicolumn{2}{c|}{\textbf{Apex}} &
        \multicolumn{2}{c|}{\textbf{Urban}} & \multirow{2}{*}{Avg.} & \multirow{2}{*}{Rank} \\
        \cline{2-9}
        & \textbf{SAD $\downarrow$} & \textbf{NMSE $\downarrow$} & \textbf{SAD $\downarrow$} & \textbf{NMSE $\downarrow$} & \textbf{SAD $\downarrow$} & \textbf{NMSE $\downarrow$} & \textbf{SAD $\downarrow$} & \textbf{NMSE $\downarrow$} & & \\
        \Xhline{4\arrayrulewidth}
        Bilinear & 0.036 & 0.078 & 0.114 & 0.258 & \textcolor{myorange}{\underline{0.117}} & 0.328 & \textcolor{myblue}{\textbf{0.093}} & 0.615 & 0.205 & 4 \\
        \hline
        Deconvolution & 0.035 & \textcolor{myorange}{\underline{0.044}} & \textcolor{myorange}{\underline{0.099}} & \textcolor{myblue}{\textbf{0.100}} & \textcolor{myblue}{\textbf{0.115}} & \textcolor{myorange}{\underline{0.295}} & 0.168 & 0.176 & \textcolor{myorange}{\underline{0.129}} & \textcolor{myblue}{\textbf{1}} \\
        \hline
        Pixel Shuffle \cite{Shi_2016} & \textcolor{myblue}{\textbf{0.023}} & 0.213 & 0.144 & 0.506 & 0.528 & 0.934 & \textcolor{myorange}{\underline{0.098}} & 0.974 & 0.428 & 5 \\
        \hline
        FiLM layer \cite{Perez_2018} & 0.038 & \textcolor{myblue}{\textbf{0.041}} & 0.120 & 0.138 & 0.119 & \textcolor{myblue}{\textbf{0.141}} & 0.191 & \textcolor{myorange}{\underline{0.198}} & \textcolor{myblue}{\textbf{0.123}} & \textcolor{myorange}{\underline{2}} \\
        \hline
        NAF \cite{Chambon_2025} & 0.903 & 0.124 & 0.658 & 0.361 & 0.423 & 0.800 & 0.584 & 0.617 & 0.559 & 6 \\
        \hline
        FeatUp \cite{Fu_2024} & 0.922 & 0.178 & 0.774 & 0.510 & 0.538 & 0.369 & 0.536 & 0.888 & 0.589 & 7 \\
        \hline
        \textbf{Linear Mapping} \cite{Ghosh_2022} & \textcolor{myorange}{\underline{0.029}} & 0.046 & \textcolor{myblue}{\textbf{0.084}} & \textcolor{myorange}{\underline{0.111}} & 0.294 & 0.418 & 0.157 & \textcolor{myblue}{\textbf{0.173}} & 0.164 & 3 \\
        \hline
    \end{tabular}
    \caption{Comparison of feature resolution augmentation methods on \textbf{Panopticon} features with a \underline{single training run}}
    \label{tab:upsampling1-panopticon}
\end{table*}
\begin{table*}
    \centering\renewcommand{\arraystretch}{1.2}
    \begin{tabular}{|l|c|c|c|c|c|c|c|c|c|c|}
        \hline
        \multirow{2}{*}{} &
        \multicolumn{2}{c|}{\textbf{Samson}} &
        \multicolumn{2}{c|}{\textbf{Jasper}} &
        \multicolumn{2}{c|}{\textbf{Apex}} &
        \multicolumn{2}{c|}{\textbf{Urban}} & \multirow{2}{*}{Avg.} & \multirow{2}{*}{Rank} \\
        \cline{2-9}
        & \textbf{SAD $\downarrow$} & \textbf{NMSE $\downarrow$} & \textbf{SAD $\downarrow$} & \textbf{NMSE $\downarrow$} & \textbf{SAD $\downarrow$} & \textbf{NMSE $\downarrow$} & \textbf{SAD $\downarrow$} & \textbf{NMSE $\downarrow$} & & \\
        \Xhline{4\arrayrulewidth}
        Bilinear ($\times 15$) & 0.036 & 0.072 & 0.103 & 0.224 & 0.250 & 0.453 & \textcolor{myorange}{\underline{0.102}} & 0.539 & 0.222 & 4 \\
        \hline
        Deconvolution ($\times 15$) & 0.035 & \textcolor{myblue}{\textbf{0.041}} & \textcolor{myorange}{\underline{0.098}} & \textcolor{myorange}{\underline{0.098}} & \textcolor{myorange}{\underline{0.113}} & 0.200 & 0.169 & 0.168 & \textcolor{myorange}{\underline{0.115}} & \textcolor{myorange}{\underline{2}} \\
        \hline
        Pixel Shuffle ($\times 15$) & \textcolor{myblue}{\textbf{0.022}} & 0.091 & 0.146 & 0.255 & 0.579 & 0.680 & \textcolor{myblue}{\textbf{0.096}} & 0.598 & 0.308 & 5 \\
        \hline
        FiLM layer ($\times 15$) & 0.039 & \textcolor{myblue}{\textbf{0.041}} & 0.114 & 0.131 & \textcolor{myblue}{\textbf{0.112}} & \textcolor{myorange}{\underline{0.168}} & 0.185 & 0.182 & 0.122 & 3 \\
        \hline
        NAF ($\times 15$) & 0.842 & 0.118 & 0.619 & 0.350 & 0.372 & 0.485 & 0.548 & 0.542 & 0.485 & 6 \\
        \hline
        FeatUp ($\times 15$) & 0.911 & 0.319 & 0.745 & 0.693 & 0.520 & 0.347 & 0.532 & 0.759 & 0.603 & 7 \\
        \hline
        \textbf{Linear Mapping} ($\times 15$) & \textcolor{myorange}{\underline{0.029}} & \textcolor{myorange}{\underline{0.042}} & \textcolor{myblue}{\textbf{0.087}} & \textcolor{myblue}{\textbf{0.078}} & 0.168 & \textcolor{myblue}{\textbf{0.166}} & 0.154 & \textcolor{myblue}{\textbf{0.148}} & \textcolor{myblue}{\textbf{0.109}} & \textcolor{myblue}{\textbf{1}} \\
        \hline
    \end{tabular}
    \caption{Comparison of feature resolution augmentation methods on \textbf{Panopticon} features with a \underline{15-training ensemble}}
    \label{tab:upsampling15-panopticon}
\end{table*}
\begin{table*}
    \centering\renewcommand{\arraystretch}{1.2}
    \begin{tabular}{|l|c|c|c|c|c|c|c|c|c|c|}
        \hline
        \multirow{2}{*}{} &
        \multicolumn{2}{c|}{\textbf{Samson}} &
        \multicolumn{2}{c|}{\textbf{Jasper}} &
        \multicolumn{2}{c|}{\textbf{Apex}} &
        \multicolumn{2}{c|}{\textbf{Urban}} & \multirow{2}{*}{Avg.} & \multirow{2}{*}{Rank} \\
        \cline{2-9}
        & \textbf{SAD $\downarrow$} & \textbf{NMSE $\downarrow$} & \textbf{SAD $\downarrow$} & \textbf{NMSE $\downarrow$} & \textbf{SAD $\downarrow$} & \textbf{NMSE $\downarrow$} & \textbf{SAD $\downarrow$} & \textbf{NMSE $\downarrow$} & & \\
        \Xhline{4\arrayrulewidth}
        Bilinear & 0.035 & 0.07 & 0.105 & 0.242 & 0.116 & 0.307 & \textcolor{myorange}{\underline{0.105}} & 0.577 & 0.195 & 4 \\
        \hline
        Deconvolution & \textcolor{myorange}{\underline{0.032}} & 0.056 & \textcolor{myblue}{\textbf{0.077}} & \textcolor{myblue}{\textbf{0.092}} & \textcolor{myorange}{\underline{0.110}} & \textcolor{myblue}{\textbf{0.144}} & 0.155 & 0.333 & \textcolor{myorange}{\underline{0.125}} & 3 \\
        \hline
        Pixel Shuffle \cite{Shi_2016} & \textcolor{myorange}{\underline{0.032}} & 0.293 & 0.201 & 0.566 & 0.559 & 0.873 & \textcolor{myblue}{\textbf{0.103}} & 0.941 & 0.446 & 5 \\
        \hline
        FiLM layer \cite{Perez_2018} & 0.041 & \textcolor{myblue}{\textbf{0.044}} & 0.121 & 0.147 & \textcolor{myblue}{\textbf{0.103}} & \textcolor{myorange}{\underline{0.146}} & 0.178 & \textcolor{myorange}{\underline{0.190}} & \textcolor{myblue}{\textbf{0.121}} & \textcolor{myblue}{\textbf{1}} \\
        \hline
        NAF \cite{Chambon_2025} & 0.864 & 0.113 & 0.659 & 0.354 & 0.392 & 0.518 & 0.567 & 0.691 & 0.520 & 6 \\
        \hline
        FeatUp \cite{Fu_2024} & 0.920 & 0.148 & 0.796 & 0.659 & 0.549 & 0.355 & 0.559 & 0.936 & 0.615 & 7 \\
        \hline
        \textbf{Linear Mapping} \cite{Ghosh_2022} & \textcolor{myblue}{\textbf{0.028}} & \textcolor{myorange}{\underline{0.047}} & \textcolor{myorange}{\underline{0.080}} & \textcolor{myorange}{\underline{0.102}} & 0.199 & 0.270 & 0.142 & \textcolor{myblue}{\textbf{0.169}} & 0.130 & \textcolor{myorange}{\underline{2}} \\
        \hline
    \end{tabular}
    \caption{Comparison of feature resolution augmentation methods on \textbf{SpecAware} features with a \underline{single training run}}
    \label{tab:upsampling1-specaware}
\end{table*}
\begin{table*}
    \centering\renewcommand{\arraystretch}{1.2}
    \begin{tabular}{|l|c|c|c|c|c|c|c|c|c|c|}
        \hline
        \multirow{2}{*}{} &
        \multicolumn{2}{c|}{\textbf{Samson}} &
        \multicolumn{2}{c|}{\textbf{Jasper}} &
        \multicolumn{2}{c|}{\textbf{Apex}} &
        \multicolumn{2}{c|}{\textbf{Urban}} & \multirow{2}{*}{Avg.} & \multirow{2}{*}{Rank} \\
        \cline{2-9}
        & \textbf{SAD $\downarrow$} & \textbf{NMSE $\downarrow$} & \textbf{SAD $\downarrow$} & \textbf{NMSE $\downarrow$} & \textbf{SAD $\downarrow$} & \textbf{NMSE $\downarrow$} & \textbf{SAD $\downarrow$} & \textbf{NMSE $\downarrow$} & & \\
        \Xhline{4\arrayrulewidth}
        Bilinear ($\times 15$) & 0.036 & 0.065 & 0.098 & 0.220 & 0.197 & 0.426 & \textcolor{myorange}{\underline{0.103}} & 0.540 & 0.211 & 4 \\
        \hline
        Deconvolution ($\times 15$) & 0.033 & 0.048 & \textcolor{myblue}{\textbf{0.071}} & \textcolor{myorange}{\underline{0.078}} & \textcolor{myblue}{\textbf{0.103}} & \textcolor{myblue}{\textbf{0.134}} & 0.158 & 0.301 & \textcolor{myorange}{\underline{0.116}} & \textcolor{myorange}{\underline{2}} \\
        \hline
        Pixel Shuffle ($\times 15$) & \textcolor{myblue}{\textbf{0.021}} & 0.088 & 0.194 & 0.298 & 0.554 & 0.607 & \textcolor{myblue}{\textbf{0.101}} & 0.554 & 0.302 & 5 \\
        \hline
        FiLM layer ($\times 15$) & 0.040 & \textcolor{myorange}{\underline{0.042}} & 0.112 & 0.142 & \textcolor{myorange}{\underline{0.115}} & \textcolor{myorange}{\underline{0.162}} & 0.187 &\textcolor{myorange}{\underline{ 0.189}} & 0.124 & 3 \\
        \hline
        NAF ($\times 15$) & 0.778 & 0.110 & 0.630 & 0.306 & 0.353 & 0.454 & 0.600 & 0.610 & 0.480 & 6 \\
        \hline
        FeatUp ($\times 15$) & 0.822 & 0.164 & 0.743 & 0.625 & 0.519 & 0.286 & 0.557 & 0.850 & 0.571 & 7 \\
        \hline
        \textbf{Linear Mapping} ($\times 15$) & \textcolor{myorange}{\underline{0.029}} & \textcolor{myblue}{\textbf{0.040}} & \textcolor{myorange}{\underline{0.077}} & \textcolor{myblue}{\textbf{0.060}} & 0.157 & 0.173 & 0.147 & \textcolor{myblue}{\textbf{0.149}} & \textcolor{myblue}{\textbf{0.104}} & \textcolor{myblue}{\textbf{1}} \\
        \hline
    \end{tabular}
    \caption{Comparison of feature resolution augmentation methods on \textbf{SpecAware} features with a \underline{15-training ensemble}}
    \label{tab:upsampling15-specaware}
\end{table*}

Tables \ref{tab:upsampling1-panopticon}, \ref{tab:upsampling15-panopticon}, \ref{tab:upsampling1-specaware} and \ref{tab:upsampling15-specaware} detail the SAD and NMSE obtained by the unmixing networks using all tested upsampling methods, with Panopticon and SpecAware as the HFMs. We display the results with both a single training ($n=1$) and the ensemble method ($n=15$). We also show the ranking of each method, using the method described in \autoref{sec:setup}, as well as an average of the metrics over all datasets for easier comparison (\textit{Avg.}). These tables show that the FiLM layer outperforms the linear mapping when running only one training run (without the ensembling method). We suggest that under limited computational resources, a single training of the unmixing framework using the FiLM layer will produce the best results. However, if more resources are available, an ensembling of 15 trainings using the linear mapping layer will result in even better performance.

\section{Additional ablation study: role of the HFMs' features}

\begin{table*}
    \centering
    \renewcommand{\arraystretch}{1.2}
    \begin{tabular}{|l|c|c|c|c|c|c|c|c|}
    \hline
    \multirow{2}{*}{} &
    \multicolumn{2}{c|}{\textbf{Samson}} &
    \multicolumn{2}{c|}{\textbf{Jasper}} &
    \multicolumn{2}{c|}{\textbf{Apex}} &
    \multicolumn{2}{c|}{\textbf{Urban}} \\
    \cline{2-9}
    & \textbf{SAD $\downarrow$} & \textbf{NMSE $\downarrow$} & \textbf{SAD $\downarrow$} & \textbf{NMSE $\downarrow$} & \textbf{SAD $\downarrow$} & \textbf{NMSE $\downarrow$} & \textbf{SAD $\downarrow$} & \textbf{NMSE $\downarrow$} \\
    \Xhline{4\arrayrulewidth}
        Noise & \textcolor{myorange}{\underline{0.041 ± 0}} & 0.056 ± 0.02 & 0.132 ± 0.04 & \textcolor{myorange}{\underline{0.133 ± 0.02}} & \textcolor{myblue}{\textbf{0.103 ± 0.01}} & \textcolor{myorange}{\underline{0.184 ± 0.06}} & 0.195 ± 0.01 & \textcolor{myorange}{\underline{0.205 ± 0.01}} \\
        \hline
        Downsampled HSI & 0.046 ± 0.01 & \textcolor{myorange}{\underline{0.051 ± 0.01}} & \textcolor{myorange}{\underline{0.129 ± 0}} & 0.172 ± 0.01 & \textcolor{myorange}{\underline{0.169 ± 0.16}} & \textcolor{myblue}{\textbf{0.166 ± 0.08}} & \textcolor{myorange}{\underline{0.192 ± 0.01}} & 0.228 ± 0.01 \\
        \hline
        \textbf{HFM features} & \textcolor{myblue}{\textbf{0.028 ± 0}} & \textcolor{myblue}{\textbf{0.043 ± 0}} & \textcolor{myblue}{\textbf{0.080 ± 0}} & \textcolor{myblue}{\textbf{0.099 ± 0.03}} & 0.172 ± 0.17 & 0.255 ± 0.19 & \textcolor{myblue}{\textbf{0.149 ± 0.02}} & \textcolor{myblue}{\textbf{0.168 ± 0.01}} \\
        \hline
    \end{tabular}
    \caption{HFM features ablation study}
    \label{tab:features_ablation}
\end{table*}
To assess the importance of the HFMs as feature encoders, we replaced the low-resolution feature input of the linear mapping layer with 1) random noise and 2) a downsampled version of the input HSI. We observed the impact on the quality of the unmixing, using SpecAware as the feature extractor. The results are reported in \autoref{tab:features_ablation} and show that using the features provided by the HFM enables a significant improvement in the quality of the unmixing.

\section{Further Tests on the Resolution Augmentation Architecture}

\begin{figure}[h!]
    \centering
    \includegraphics[width=\linewidth]{ 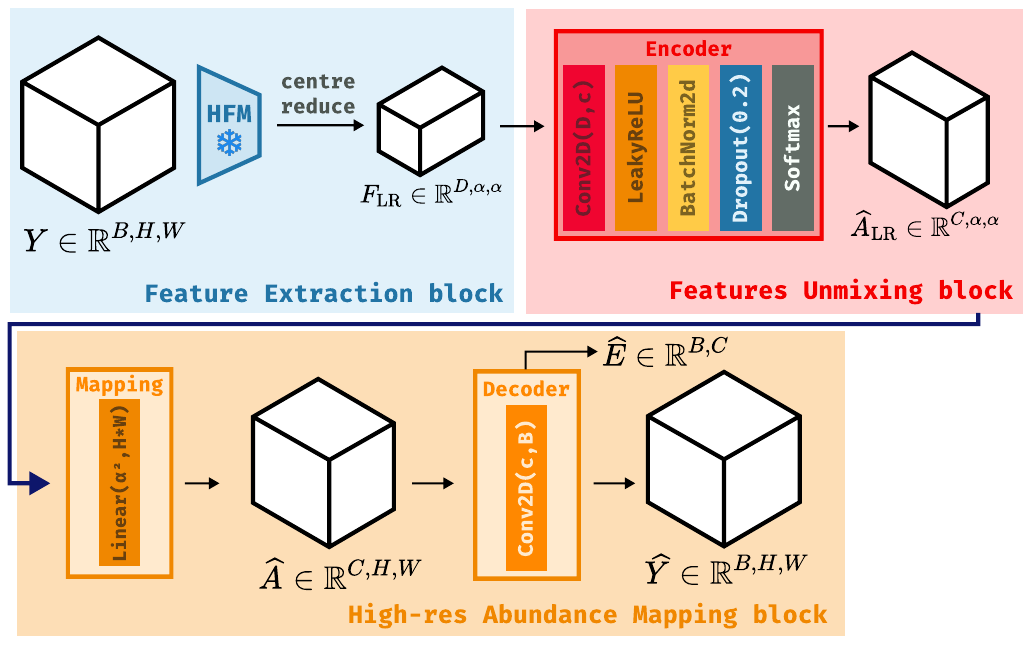}
    \caption{Reversed pipeline: we first unmix low-resolution features and then upsample the estimated abundances}
    \label{fig:unmixing-reversed}
\end{figure}

To analyse the performance of the proposed resolution augmentation architecture, we tried to permute the mapping and unmixing blocks. In detail, we first unmix the low-resolution features using the proposed abundance estimation CNN. A low-resolution estimation of the abundances is obtained and mapped back to high-resolution using the linear layer. It is then forwarded to the decoder to reconstruct the input HSI (\autoref{fig:unmixing-reversed}).
\autoref{tab:unmixing-order-comparison} contains the results obtained with both methods and shows that the proposed method: first projects the features to high resolution and then unmixes, achieves the best performance.

\begin{table*}[]
    \centering
    \renewcommand{\arraystretch}{1.2}
    \addtolength{\tabcolsep}{-4pt}
    \begin{tabular}{|l|c|c|c|c|c|c|c|c|}
        \hline
        \multirow{2}{*}{} &
        \multicolumn{2}{c|}{\textbf{Samson}} &
        \multicolumn{2}{c|}{\textbf{Jasper}} &
        \multicolumn{2}{c|}{\textbf{Apex}} &
        \multicolumn{2}{c|}{\textbf{Urban}} \\
        \cline{2-9}
        & \textbf{SAD $\downarrow$} & \textbf{NMSE $\downarrow$} & \textbf{SAD $\downarrow$} & \textbf{NMSE $\downarrow$} & \textbf{SAD $\downarrow$} & \textbf{NMSE $\downarrow$} & \textbf{SAD $\downarrow$} & \textbf{NMSE $\downarrow$} \\
        \Xhline{4\arrayrulewidth}
        $\widehat{A}_{\text{LR}}$ unmixing \autoref{fig:unmixing-reversed} & \textcolor{myorange}{\underline{0.035 ± 0}} & 0.202 ± 0.02 & 0.098 ± 0.013 & 0.101 ± 0.09 & \textcolor{myblue}{\textbf{0.160 ± 0.01}} & \textcolor{myorange}{\underline{0.235 ± 0.01}} & 0.323 ± 0.01 & \textcolor{myorange}{\underline{0.453 ± 0.11}} \\
        \hline
        \textbf{Proposed unmixing} \autoref{fig:fm-unmixing} & \textcolor{myblue}{\textbf{0.028 ± 0}} & \textcolor{myblue}{\textbf{0.043 ± 0}} & \textcolor{myorange}{\underline{0.080 ± 0}} & \textcolor{myblue}{\textbf{0.099 ± 0.03}} & 0.172 ± 0.17 & 0.255 ± 0.19 & \textcolor{myblue}{\textbf{0.149 ± 0.02}} & \textcolor{myblue}{\textbf{0.168 ± 0.01}} \\
        \hline
    \end{tabular}
    \caption{Comparison of the different unmixing pipelines}
    \label{tab:unmixing-order-comparison}
\end{table*}

\section{Visual comparison of methods on other datasets}

\begin{figure*}[h]
    \centering
    \includegraphics[width=\linewidth]{ 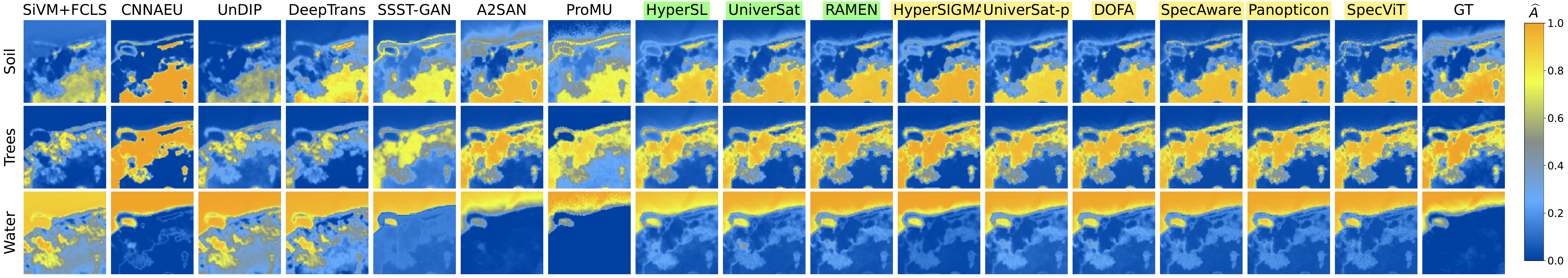}
    \caption{Comparison of all method's estimated $\widehat{A}$ on the Samson dataset}
  \label{fig:a-comparison-samson}
\end{figure*}

\begin{figure*}[h]
    \centering
    \includegraphics[width=\linewidth]{ 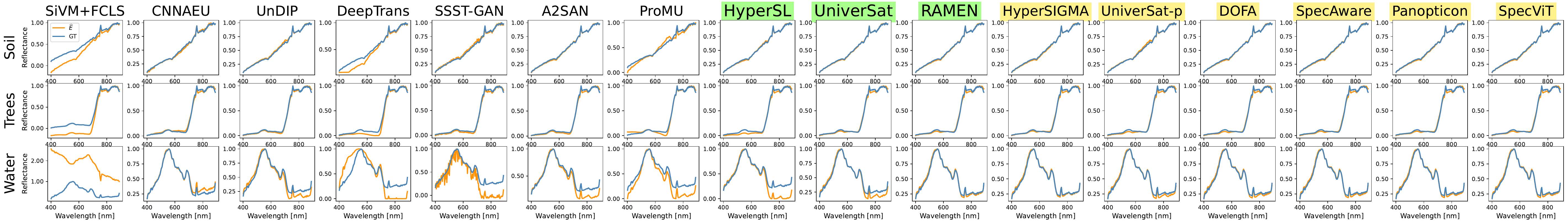}
    \caption{Comparison of all method's estimated $\widehat{E}$ on the Samson dataset}
  \label{fig:e-comparison-samson}
\end{figure*}

\begin{figure*}[h]
    \centering
    \includegraphics[width=\linewidth]{ 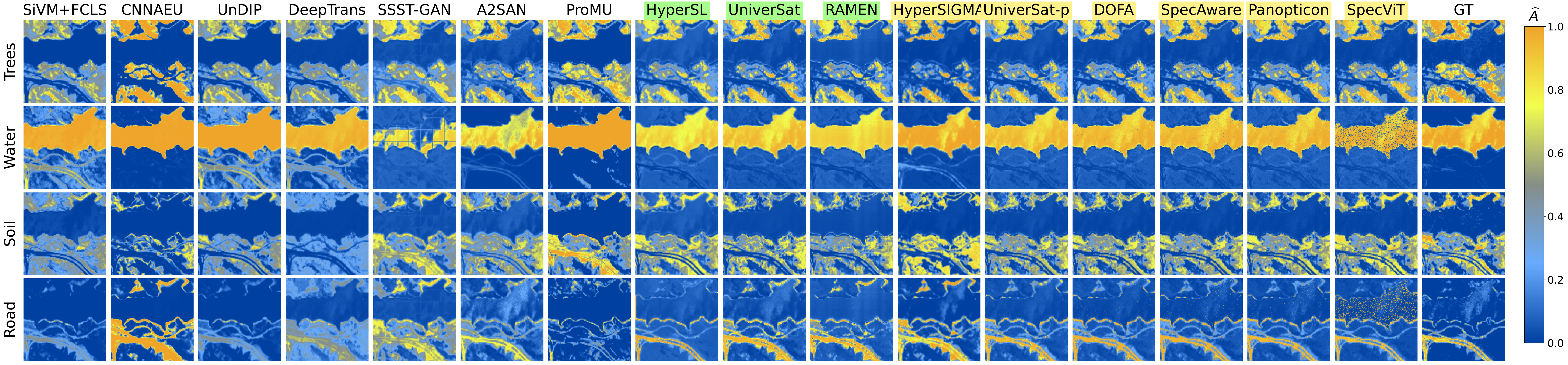}
    \caption{Comparison of all method's estimated $\widehat{A}$ on the Jasper dataset}
  \label{fig:a-comparison-jasper}
\end{figure*}

\begin{figure*}[h]
    \centering
    \includegraphics[width=\linewidth]{ 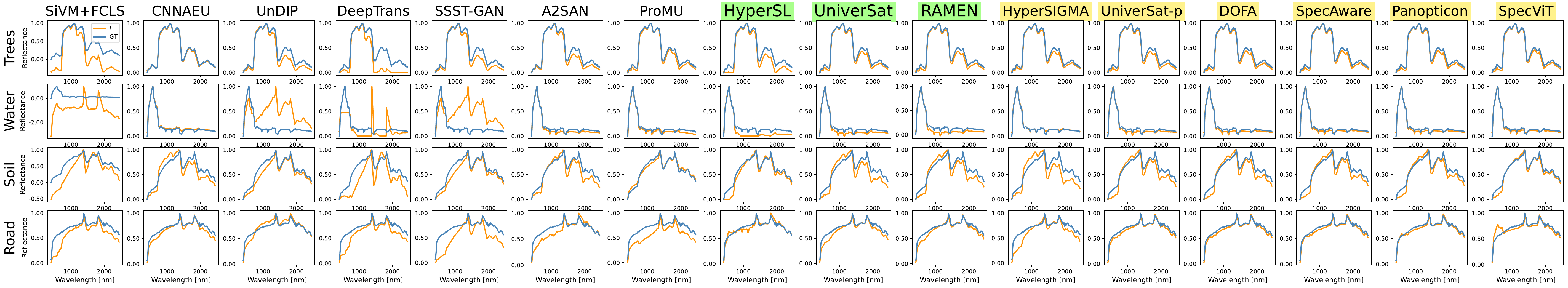}
    \caption{Comparison of all method's estimated $\widehat{E}$ on the Jasper dataset}
  \label{fig:e-comparison-jasper}
\end{figure*}

\begin{figure*}[h]
    \centering
    \includegraphics[width=\linewidth]{ 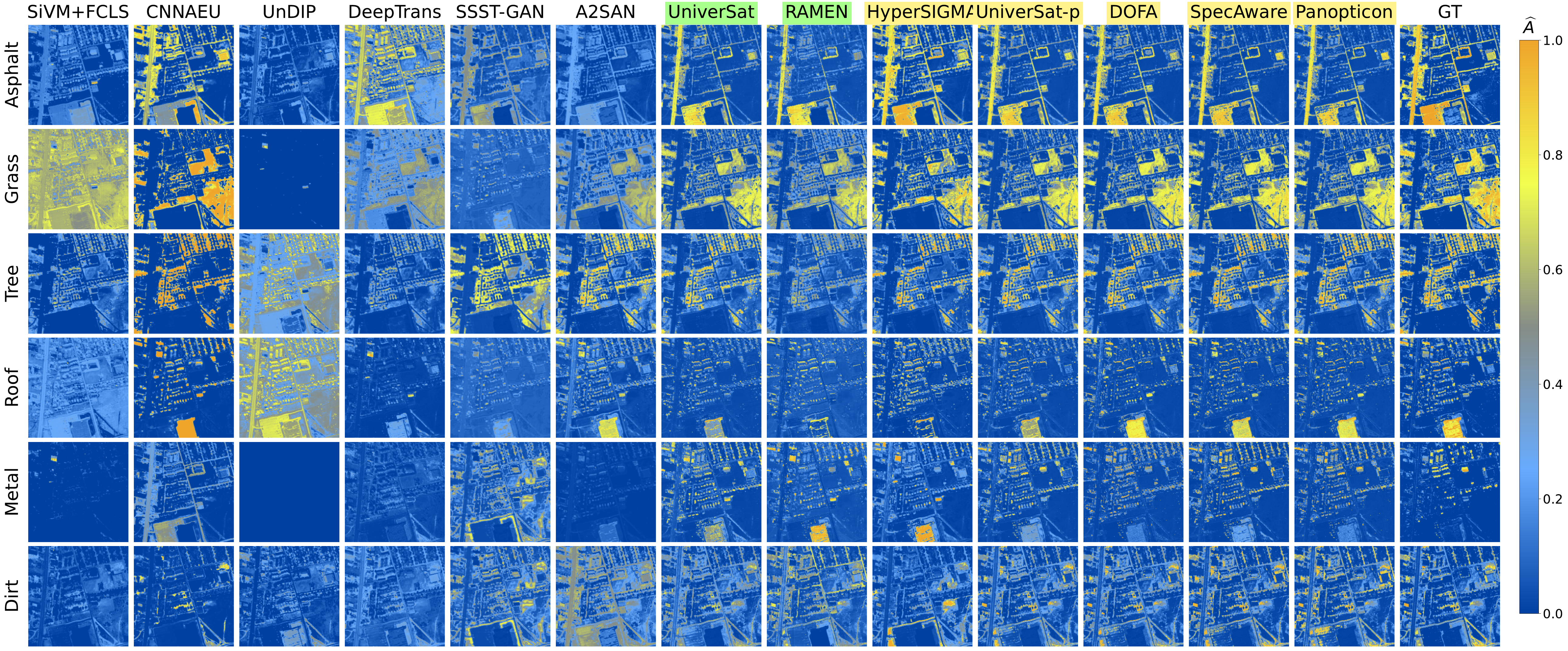}
    \caption{Comparison of all method's estimated $\widehat{A}$ on the Urban dataset}
  \label{fig:a-comparison-urban}
\end{figure*}

\begin{figure*}[h]
    \centering
    \includegraphics[width=\linewidth]{ 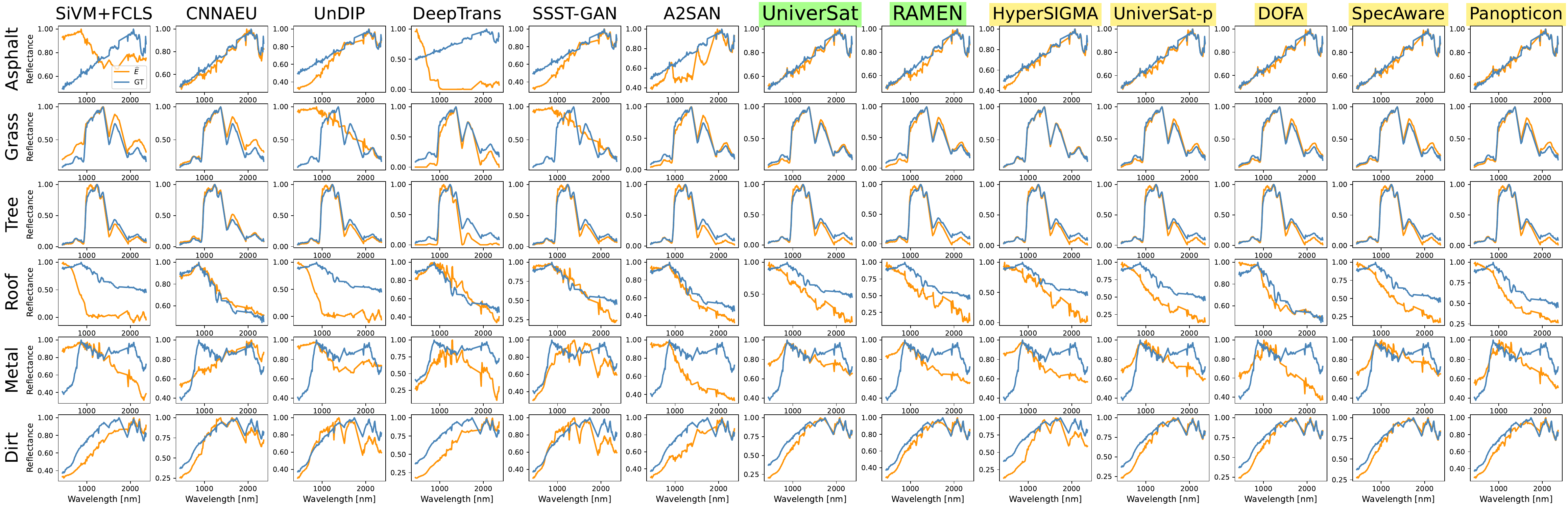}
    \caption{Comparison of all method's estimated $\widehat{E}$ on the Urban dataset}
  \label{fig:e-comparison-urban}
\end{figure*}

Figures \ref{fig:a-comparison-samson} to \ref{fig:e-comparison-urban} provide other visual comparisons of the HFMs using the proposed framework to the other state-of-the-art methods on the Samson, Jasper, and Urban datasets, in addition to the Apex dataset presented in the letter. We can see that HFMs surpass most state-of-the-art methods, and the ones using our proposed framework surpass HyperSIGMA and HyperSL. In particular, we can see that our method is expressive enough to capture the correct variations in intensity in the water and the road in the Jasper dataset, for example. In the same way, we can see that the HFMs are able to extract fine details in the road and the soil in the Urban dataset.

\begin{figure*}[h]
    \centering
    \includegraphics[width=\linewidth]{ 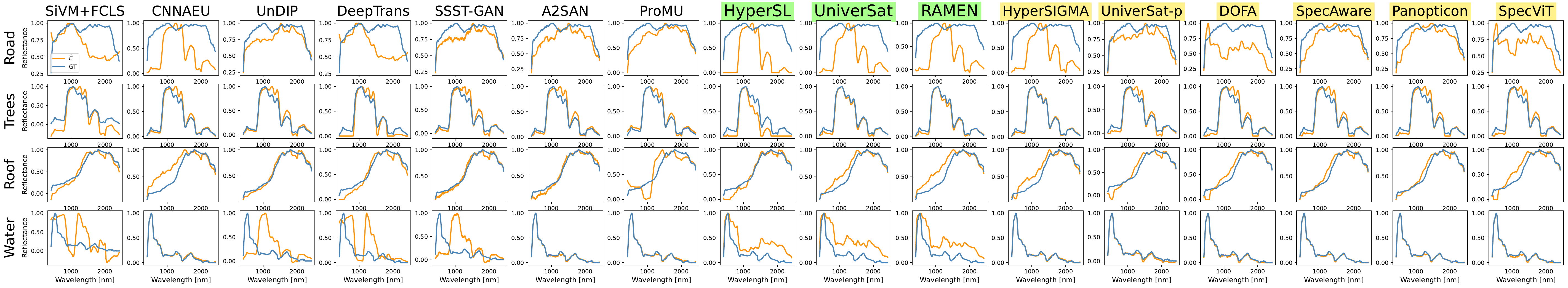}
    \caption{Comparison of all method's estimated $\widehat{E}$ on the Apex dataset}
  \label{fig:e-comparison-apex}
\end{figure*}

\end{document}